\def\year{2022}\relax
\documentclass[letterpaper]{article} % DO NOT CHANGE THIS
\usepackage{aaai22}  % DO NOT CHANGE THIS
\usepackage{times}  % DO NOT CHANGE THIS
\usepackage{helvet}  % DO NOT CHANGE THIS
\usepackage{courier}  % DO NOT CHANGE THIS
\usepackage[hyphens]{url}  % DO NOT CHANGE THIS
\usepackage{graphicx} % DO NOT CHANGE THIS
\usepackage{natbib}  % DO NOT CHANGE THIS AND DO NOT ADD ANY OPTIONS TO IT
\usepackage{caption} % DO NOT CHANGE THIS AND DO NOT ADD ANY OPTIONS TO IT
\DeclareCaptionStyle{ruled}{labelfont=normalfont,labelsep=colon,strut=off} % DO NOT CHANGE THIS
\usepackage{algorithm}
\usepackage{algorithmic}

\usepackage{newfloat}
\usepackage{listings}
\floatstyle{ruled}
\newfloat{listing}{tb}{lst}{}
\floatname{listing}{Listing}
\usepackage{pgfgantt}
\usepackage{amsmath,amssymb}
\usepackage{tikz}
\usetikzlibrary{shapes.geometric, arrows}
\usepackage[normalem]{ulem}
\usepackage{caption}
\usepackage{url}
\newcommand{\rom}[1]{\uppercase\expandafter{\romannumeral #1\relax}}

\usepackage{fancyhdr}
\usetikzlibrary{positioning, fit, arrows.meta, shapes}
\usepackage{subcaption}
\usepackage[toc,page]{appendix}

\title{D-Flow: A Real Time Spatial Temporal Model for Target Area Segmentation}
\providecommand{\keywords}[1]
{
  \small	
  \textbf{\textit{Keywords---}} #1
}

\usepackage{bibentry}
\begin{document}

\author{
Wentao Lu,
Claude Sammut
}
\affiliations{School of Computer Science and Engineering\\
The University of New South Wales\\
\{wentao.lu, c.sammut\}@unsw.edu.au}

\maketitle

\begin{abstract}
Semantic segmentation has attracted  a large amount of attention in recent years. In robotics, segmentation can be used to identify a region of interest, or \emph{target area}. For example, in the RoboCup Standard Platform League (SPL), segmentation separates the soccer field from the background and from players on the field. For satellite or vehicle applications, it is often necessary to find certain regions such as roads, bodies of water or kinds of terrain. In this paper, we propose a novel approach to real-time target area segmentation based on a newly designed spatial temporal network. The method operates under domain constraints defined by both the robot's hardware and its operating environment . The proposed network is able to run in real-time, working within the constraints of limited run time and computing power. This work is compared against other real time segmentation methods on a dataset generated by a Nao V6 humanoid robot simulating the RoboCup SPL competition. In this case, the target area is defined as the artificial grass field. The method is also tested on a maritime dataset collected by a moving vessel, where the aim is to separate the ocean region from the rest of the image. This dataset demonstrates that the proposed model can generalise to a variety of vision problems.
\end{abstract}

\section{Introduction}

The aim of the work presented here is to efficiently segment regions of interest in a robot's vision system.
Image segmentation must work within a set of tight constraints.  The computational resources available on-board a robot are often very limited. Furthermore, conserving power is critical in mobile robot, which often leads to trade-offs in computing power, making running time a major concern.Dynamic lighting conditions significantly affect the robot's vision, hence many colour-based vision segmentation methods fail.

There have been many recent proposals for real time image segmentation. However, most of them focus on static semantic segmentation and cannot function effectively under the above constraints. Because of the focus on static images, they do not take advantage of temporal information in a video stream.

\begin{figure*}
\centering

\tikzstyle{convrecurrent} = [rectangle, minimum width=1cm, minimum height=0.5cm, text centered, draw=black, fill=orange!30]
\tikzstyle{non-convrecurrent} = [rectangle, minimum width=1cm, minimum height=0.5cm, text centered, draw=black, fill=gray!30]
\tikzstyle{arrow} = [thick,->,>=stealth]

\begin{tikzpicture}[node distance=0.5cm]
\node (t-k) [non-convrecurrent] {$RGB_{t-k}$};
\node (t-k+1) [non-convrecurrent, right of=t-k, node distance=3cm] {$RGB_{t-k+1}$};
\node (t) [non-convrecurrent, right of=t-k+1, node distance=5cm] {$RGB_{t}$};

\node at ($(t-k+1)!.5!(t)$) {\ldots};

\node (rgb1) [convrecurrent, above of=t-k, node distance=1cm] {ConvMGU};
\node (rgb2) [convrecurrent, right of=rgb1, node distance=3cm] {ConvMGU};
\node (rgb3) [convrecurrent, right of=rgb2, node distance=5cm] {ConvMGU};

\node at ($(rgb2)!.5!(rgb3)$) {\ldots};

\node (rgb4) [convrecurrent, above of=rgb1, node distance=1cm] {ConvMGU};
\node (rgb5) [convrecurrent, above of=rgb2, node distance=1cm] {ConvMGU};
\node (rgb6) [convrecurrent, above of=rgb3, node distance=1cm] {ConvMGU};

\node at ($(rgb5)!.5!(rgb6)$) {\ldots};

\node (yuvinput1) [non-convrecurrent, below of=t-k, node distance=1cm] {$YUV_{t-k}$};
\node (yuvinput2) [non-convrecurrent, right of=yuvinput1, node distance=3cm] {$YUV_{t-k+1}$};
\node (yuvinput3) [non-convrecurrent, right of=yuvinput2, node distance=5cm] {$YUV_{t}$};

\node at ($(yuvinput2)!.5!(yuvinput3)$) {\ldots};

\node (yuv1) [convrecurrent, below of=yuvinput1, node distance=1cm] {ConvMGU};
\node (yuv2) [convrecurrent, right of=yuv1, node distance=3cm] {ConvMGU};
\node (yuv3) [convrecurrent, right of=yuv2, node distance=5cm] {ConvMGU};

\node at ($(yuv2)!.5!(yuv3)$) {\ldots};

\node (yuv4) [convrecurrent, below of=yuv1, node distance=1cm] {ConvMGU};
\node (yuv5) [convrecurrent, below of=yuv2, node distance=1cm] {ConvMGU};
\node (yuv6) [convrecurrent, below of=yuv3, node distance=1cm] {ConvMGU};

\node at ($(yuv5)!.5!(yuv6)$) {\ldots};

\node (add) [non-convrecurrent, below right =0.1cm and 1.9cm of t] {Add};
\node (conv) [non-convrecurrent, right of=add, node distance=3cm]{Conv2d};
\node (out) [non-convrecurrent, below of=conv, node distance=3cm]{Output};

\draw [arrow] (rgb1) -- (rgb2);
\draw [arrow] (rgb4) -- (rgb5);
\draw [arrow] (yuv1) -- (yuv2);
\draw [arrow] (yuv4) -- (yuv5);

\draw [arrow] (t-k) -- (rgb1);
\draw [arrow] (t-k+1) -- (rgb2);
\draw [arrow] (t) -- (rgb3);
\draw [arrow] (rgb1) -- (rgb4);
\draw [arrow] (rgb2) -- (rgb5);
\draw [arrow] (rgb3) -- (rgb6);
\draw [->, thick] (rgb2) -- ++(2cm, 0);
\draw [->, thick] (rgb5) -- ++(2cm, 0);
\draw [->, thick] (rgb2)++(3cm, 0) -- ++(0.9cm, 0);
\draw [->, thick] (rgb5)++(3cm, 0) -- ++(0.9cm, 0);

\draw [arrow] (t-k) -- (yuvinput1);
\draw [arrow] (t-k+1) -- (yuvinput2);
\draw [arrow] (t) -- (yuvinput3);
\draw [arrow] (yuvinput1) -- (yuv1);
\draw [arrow] (yuvinput2) -- (yuv2);
\draw [arrow] (yuvinput3) -- (yuv3);

\draw [arrow] (yuv1) -- (yuv4);
\draw [arrow] (yuv2) -- (yuv5);
\draw [arrow] (yuv3) -- (yuv6);

\draw [->, thick] (yuv2) -- ++(2cm, 0);
\draw [->, thick] (yuv5) -- ++(2cm, 0);
\draw [->, thick] (yuv2)++(3cm, 0) -- ++(0.9cm, 0);
\draw [->, thick] (yuv5)++(3cm, 0) -- ++(0.9cm, 0);

\draw [-, thick] (rgb6) -- ++(3cm, 0);
\draw [arrow] (rgb6)++(3cm, 0) -- (add);

\draw [-, thick] (yuv6) -- ++(3cm, 0);
\draw [arrow] (yuv6)++(3cm, 0) -- (add);
\draw [arrow] (add) -- (conv);
\draw [arrow] (conv) -- (out);

\end{tikzpicture}
\caption{D-Flow Network Overview} \label{fig:1}
\end{figure*}
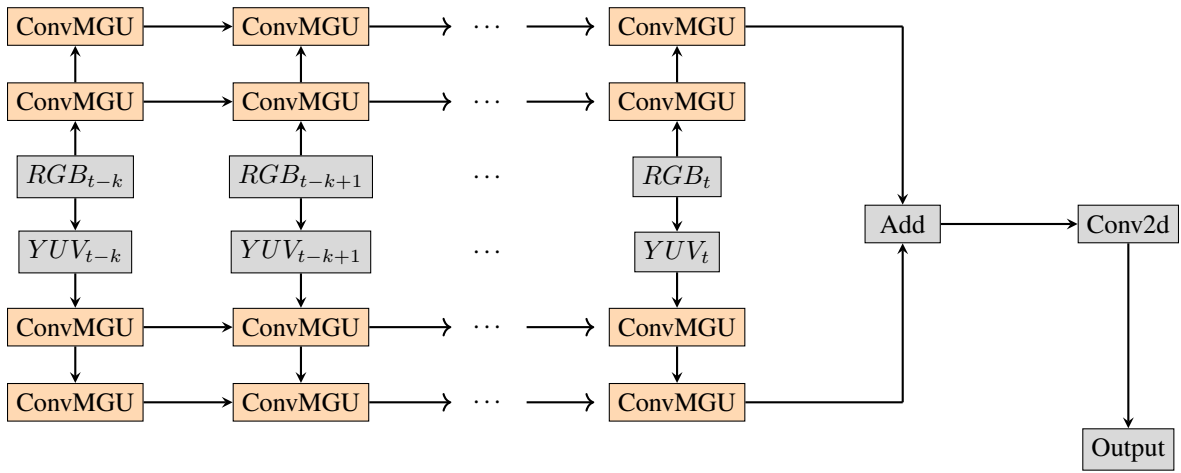

Assuming that robot vision can be treated as processing a sequence of images, the model proposed in this paper tries to learn from both spatial and temporal information based on a number of consecutive frames to predict the segmentation of region of interest (ROI) of the final frame.
Figure \ref{fig:1} presents a structural overview of the proposed model, called \emph{D-Flow} because it has two spatial temporal sub-nets to process the data flow in two colour spaces. The basic recurrent unit is a newly designed cell where we introduce convolution operations into a simplified gated recurrent unit known as the Minimal Gated Unit~\cite{Zhou2016}. This new convolutional recurrent cell can perform spatial temporal feature extraction while significantly reducing the number of parameters involved.

There are two main reasons for splitting the data flow into two separate sub-nets. First, the model should be robust to changing lighting conditions. Second, although the transform from RGB to YUV is linear and it can be done with a single flow, a single flow architecture can result in a larger number of parameters. Splitting the data into two flows acts as a heuristic to guide each flow to emphasise a particular sub task, which we describe below.

In our first experiment, the SPL robot vision dataset is obtained from a Nao V6 humanoid robot\footnote{https://www.softbankrobotics.com/emea/en/nao} and the ROI  or \emph{target area} is defined as the soccer field as per the rules of the RoboCup SPL competition. Our model and two other real time semantic segmentation methods, Fast-SCNN~\cite{DBLP:journals/corr/abs-1902-04502} and a Fully Convolutional Segmentation Network~\cite{Long_2015_CVPR}, are trained on this dataset for comparison. The results are analysed, both quantitatively and qualitatively.

The rest of this paper is organised as follows: first, we address related work, discussing other spatial temporal models and semantic segmentation methods. The design of the new model is described in the following section. We then compare both the quantitative and qualitative results from the SPL dataset with the other two baseline models along with another handcrafted methods. Also, we apply the proposed model to a Maritime dataset to demonstrate the generality of the method. We also perform a set of experiments with different colour spaces to explain why YUV and RGB are chosen in our model.

\section{Related work}

The two main components of this work are the Spatial Temporal Learning model and the Vision Segmentation method. In this section, we discuss previous work related to these two components.

\subsection{Spatial Temporal Learning Model}

\begin{figure}[!h]
	\centering
	\includegraphics[width=0.7\linewidth]{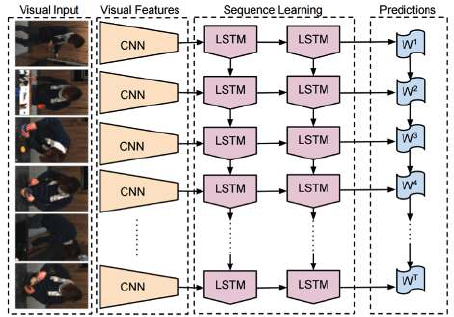}
	\caption{Long-term Recurrent Convolution Networks}
	\label{fig:LRCN}
\end{figure}

Spatial Temporal Learning models are capable of learning both spatial and temporal information based on a dataset. One method for combining Spatial and Temporal Leaning models is the introduction of both a convolution network~\cite{LeCun1999} and a recurrent network into a single model. Previous work has presented different alternatives to doing this. Donahue \emph{et al.}~\cite{Donahue_2015_CVPR} proposed a combined model where the image data are fed into the CNN layers first to extract visual features that are then processed by a sequence of LSTM~\cite{doi:10.1162/089976600300015015}\cite{doi:10.1162/neco.1997.9.8.1735} units to encode temporal features. This is shown in Figure \ref{fig:LRCN}.

\begin{figure}[!h]
	\centering
	\includegraphics[width=0.7\linewidth]{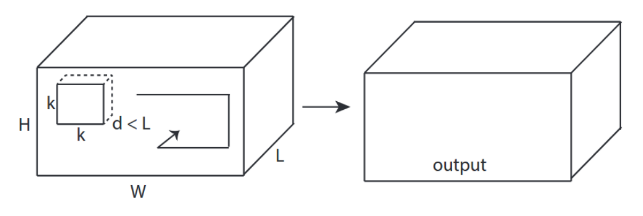}
	\caption{3D-Convolutional Neural Networks}
	\label{fig:3dcnn}
\end{figure}

Another similar approach was proposed by Tran \emph{et al.}~\cite{Tran_2015_ICCV}. The authors describe a 3D convolutional neural network, shown in Figure \ref{fig:3dcnn}, where the third dimension of each convolution kernel is a convolution operation over time. This model does not contain any recurrent units that require previous states as an extra input to the process. Therefore, one of the major advantages of this model is that its run time is faster than those models based on recurrent networks.

\begin{figure}[!h]
	\centering
	\includegraphics[width=0.7\linewidth]{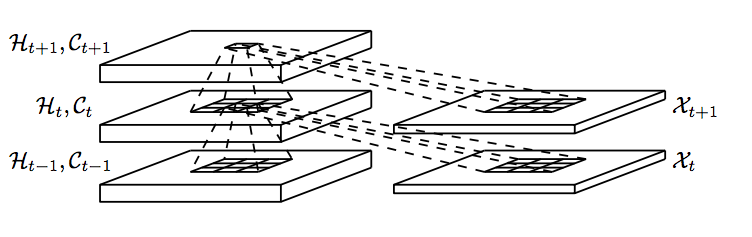}
	\caption{ConvLSTM}
	\label{fig:ConvLSTM}
\end{figure}

Figure \ref{fig:ConvLSTM} shows a novel convolutional recurrent unit along with an encoder-decoder structure proposed by Shi el al.~\cite{NIPS2015_5955}, much like the cascade architecture introduced by Donahue \emph{et al}. The main innovation of this model is that the authors have replaced the internal matrix multiplication operation with a convolution operation that allows spatial features to be extracted by the new recurrent unit. While the model is successful in solving certain problems, there is a drawback when applying it to real-time robotics problems, which is the relatively complex cell structure and large number of parameters, since the base model is a LSTM.

\subsection{Image Segmentation}

There are many recent models capable of performing image segmentation, including FCN~\cite{Long_2015_CVPR}, SegNet~\cite{DBLP:journals/corr/BadrinarayananK15} and DeepLabv3+~\cite{Chen_2018_ECCV}. They use an object recognition network, usually a deep CNN structure, as the visual feature encoder followed by a decoder network. These models demonstrate state of the art performance on some image datasets, like COCO\cite{lin2015microsoft} or PASCAL\cite{Everingham10}. However, in our case, the only semantic class in our data is a region of interest, which may have an arbitrary shape, depending on the robot's view pose. In other words, the target is not a group of objects but a group of pixels that have similar colours. Furthermore, these object recognition networks usually have a large number of parameters and very complex structures, making them computationally expensive to run in real time on limited hardware.

\section{Network Structure}

In this section, we present a new convolutional recurrent unit, called ConvMGU, and explain the design of both the unit and the D-Flow Network that utilises it.

\subsection{Convolutional Minimal Gated Unit}

\begin{figure}
\centering
\begin{tikzpicture}[
% GLOBAL CFG
font=\sf \scriptsize,
>=LaTeX,
% Styles
cell/.style={% For the main box
	rectangle, 
	rounded corners=5mm, 
	draw,
	dashed,
	thick,
},
operator/.style={%For operators like +  and  x
	circle,
	draw,
	inner sep=-0.5pt,
	minimum height =.3cm,
},
function/.style={%For functions
	ellipse,
	draw,
	inner sep=1pt
},
ct/.style={% For external inputs and outputs
	rectangle,
	draw,
	minimum width=8mm,
	minimum height=6mm,
	inner sep=1pt
},
gt/.style={% For internal inputs
	rectangle,
	draw,
	minimum width=6mm,
	minimum height=6mm,
	inner sep=1pt
},
act/.style={% For activations
	rectangle,
	draw,
	minimum width=12mm,
	minimum height=6mm,
	inner sep=1pt
},
mylabel/.style={% something new that I have learned
	font=\scriptsize\sffamily
},
ArrowC1/.style={% Arrows with rounded corners
	rounded corners=.25cm,
	thick,
},
ArrowC2/.style={% Arrows with big rounded corners
	rounded corners=.5cm,
	thick,
},
]

%Start drawing the thing...    
% Draw the cell: 
\node [cell, minimum height =4cm, minimum width=6cm] at (0,0){} ;

% Draw inputs named ibox#
\node [gt] (wf) at (-2,-0.75) {$W_f$};
\node [gt] (uf) at (-1.4,-0.75) {$U_f$};
\node [act] (sf) at (-1.7, -0.15) {$\sigma_g$};

\node [gt] (1-) at (-0.3,-0.15) {1-};
%\node [operator] (mux2) at (-0.5,0) {$\times$};
\node [operator] (dot1) at (-0.3, -0.75) {$\cdot$};
\node [operator] (dot2) at (0.5, -0.3) {$\cdot$};

\node [gt] (wh) at (1.3,-0.75) {$W_h$};
\node [gt] (uh) at (1.9,-0.75) {$U_h$};
\node [act] (sh) at (1.6, -0.15) {$\sigma_h$};

\node [operator] (add1) at (2.6, -0.3) {+};
\node [operator] (dot3) at (1.6, 0.7) {$\cdot$};

% Draw External inputs? named as basis c,h,x
\node[ct, label={[mylabel]Hidden}] (h) at (-4,-1.5) {$h_{t-j-1}$};
\node[ct, label={[mylabel]below:Input Frame j}] (x) at (-2.7,-3) {$x_{t-j}$};
\node[ct, label={}] (x1) at (-4.1,-3) {$x_{t-j-1}$};

% Draw External outputs? named as basis c2,h2,x2

% \node[ct, label={[mylabel]Next Hidden}] (h2) at (4,-1.5) {\empt{h}{t-j}};

\draw [->, ArrowC1] (h) -| (uf);
\draw [->, ArrowC1] (h) -| (dot1);
\draw [->, ArrowC1] (h) -| (dot2);

\draw [->, ArrowC1] (x) -| (wf);
\draw [->, ArrowC1] (x) -| (wh);

\draw [->, ArrowC1] (sf) |- (dot3);
\draw [->, ArrowC1]  (1-)++(0, 0.6)-| (1-);
\draw [->, ArrowC1] (dot2)++(0, 0.75) -| (dot2);
\draw [->, ArrowC1] (sh) -- (dot3);
\draw [->, ArrowC1] (dot2)++(0.15, 0) -| ++(0.3, -1.3) -| (uh);

\draw [->, ArrowC1] (dot3) -| (add1);
\draw [->, ArrowC1] (dot1)++(0.15, 0)-| ++(0.3, -1.1) -| (add1);
\draw [->, ArrowC1] (add1)++(0.15, 0) -| ++(0.3, -0.6) |- (4, -1.5);

\end{tikzpicture}
\caption{ConvMGU Unit}
\label{ConvMGU Unit}
\end{figure}
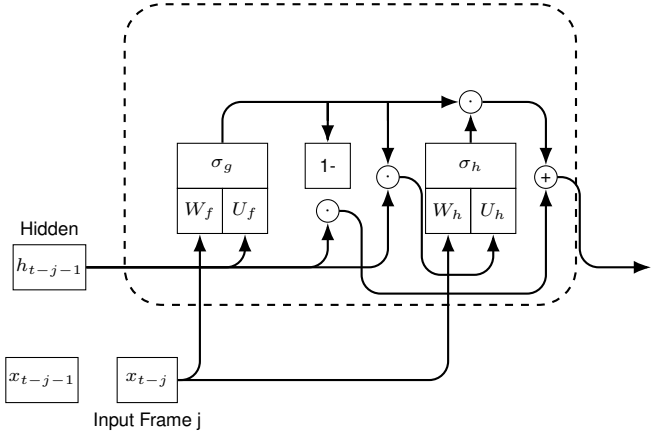

Inspired by the MGU~\cite{Zhou2016} where the authors merge two gates in the original LSTM design to reduce the number of parameters, we borrow this idea to solve a similar problem when designing the new convolution recurrent unit. Figure \ref{ConvMGU Unit} shows the infrastructure of this unit.

\begin{equation}
f_t = \sigma_g(W_{f} \ast x_t + U_{f} \ast h_{t-1} + b_f) \\
\end{equation}

\begin{equation}
h_t =  (1-f_t) \circ h_{t-1} + f_t \circ \sigma_h(W_{h}
\ast x_t + U_{h} \ast (f_t \circ h_{t-1}) + b_h)
\end{equation}

Equations 1 and 2 are modified from the original MGU paper where the internal matrix multiplications are replaced with convolution operations as denoted by $\ast$. $f_{t}$ is the output of the current step. It is used, along with the hidden state $h_{t-1}$, coming from the last step, to calculate the hidden state $h_{t}$ of the current step. These modifications introduce the ability to extract spatial features while maintaining the major advantage of the original MGU which is the relatively small number of parameters.

\begin{figure}
\centering
\begin{tikzpicture}[
% GLOBAL CFG
font=\sf \scriptsize,
>=LaTeX,
% Styles
cell/.style={% For the main box
	rectangle, 
	rounded corners=5mm, 
	draw,
	dashed,
	thick,
},
cell2/.style={% For the main box
	rectangle, 
	rounded corners=5mm, 
	draw,
	thick,
},
3dcell/.style={% For the main box
	cube, 
	rounded corners=5mm, 
	draw,
	thick,
},
operator/.style={%For operators like +  and  x
	circle,
	draw,
	inner sep=-0.5pt,
	minimum height =.3cm,
},
function/.style={%For functions
	ellipse,
	draw,
	inner sep=1pt
},
ct/.style={% For external inputs and outputs
	rectangle,
	draw,
	minimum width=8mm,
	minimum height=6mm,
	inner sep=1pt
},
gt/.style={% For internal inputs
	rectangle,
	draw,
	minimum width=6mm,
	minimum height=6mm,
	inner sep=1pt
},
act/.style={% For activations
	rectangle,
	draw,
	minimum width=12mm,
	minimum height=6mm,
	inner sep=1pt
},
mylabel/.style={% something new that I have learned
	font=\scriptsize\sffamily
},
ArrowC1/.style={% Arrows with rounded corners
	rounded corners=.25cm,
	thick,
},
ArrowC2/.style={% Arrows with big rounded corners
	rounded corners=.5cm,
	thick,
},
]

\node [cell, minimum height =4cm, minimum width=6cm] at (0,0){} ;

\node [cell2, minimum height =5cm, minimum width=2.5cm] at (-2,0){} ;

\pgfmathsetmacro{\cubex}{1}
\pgfmathsetmacro{\cubey}{1}
\pgfmathsetmacro{\cubez}{1}

\draw[label={3DCNN}] (-4,0,0) -- ++(-\cubex,0,0) -- ++(0,-\cubey,0) -- ++(\cubex,0,0) -- cycle;
\draw[] (-4,0,0) -- ++(0,0,-\cubez) -- ++(0,-\cubey,0) -- ++(0,0,\cubez) -- cycle;
\draw[] (-4,0,0) -- ++(-\cubex,0,0) -- ++(0,0,-\cubez) -- ++(\cubex,0,0) -- cycle;
\node[draw] (3d) at (-4.5, -1.3, 0) {1x1CNN};

\node[ct, label={}] (x1) at (-2,-3) {$x_{t-j}$};
\node[ct, label={}] (x2) at (0.2,-3) {$x_{t-j+1}$};
\node[ct, label={}] (x3) at (2,-3) {$x_{t}$};

\node[ct, label={}] (m11) at (-2,-1.2) {ConvMGU};
\node[ct, label={}] (m12) at (-2,0) {ConvMGU};
\node [operator] (add1) at (-2, 1) {+};

\node[ct, label={}] (m21) at (0.2,-1.2) {ConvMGU};
\node[ct, label={}] (m22) at (0.2,0) {ConvMGU};
\node [operator] (add2) at (0.2, 1) {+};

\node[ct, label={}] (m31) at (2,-1.2) {ConvMGU};
\node[ct, label={}] (m32) at (2,0) {ConvMGU};
\node [operator] (add3) at (2, 1) {+};

\draw [->, ArrowC1] (x1) -- (m11);
\draw [->, ArrowC1] (m11) -- (m12);
\draw [->, ArrowC1] (m12) -- (add1);
\draw [very thick] (x1) -- (-2,-3.8);
\draw [->, ArrowC1] (x1)++(0, -0.8) -| (3d);

\draw [->, ArrowC1] (x2) -- (m21);
\draw [->, ArrowC1] (m21) -- (m22);
\draw [->, ArrowC1] (m22) -- (add2);
\draw [very thick] (x2) -- (0.2,-3.8);
\draw [->, ArrowC1] (x2)++(0, -0.8) -| (3d);

\draw [->, ArrowC1] (x3) -- (m31);
\draw [->, ArrowC1] (m31) -- (m32);
\draw [->, ArrowC1] (m32) -- (add3);
\draw [very thick] (x3) -- (2,-3.8);
\draw [->, ArrowC1] (x3)++(0, -0.8) -| (3d);

\draw[very thick]  (-4.4, 0.15) -- (-4.4, 1.65);
\draw [->, ArrowC1] (-4.4, 1.65) -| (add1);
\draw [->, ArrowC1] (-4.4, 1.65) -| (add2);
\draw [->, ArrowC1] (-4.4, 1.65) -| (add3);

\draw [->, ArrowC1] (m11) -- (m21);

\draw [very thick] (add1) -- (-1, 1);
\draw [->, ArrowC1] (-1, 1) |- (m22);

\draw [->, thick] (add2) -- (1.2, 1);
\draw [->, thick] (add3) -- (3.5, 1);

\node at ($(m22)!.5!(m32)$) {\ldots};
\node at ($(m21)!.5!(m31)$) {\ldots};
\node at ($(x2)!.5!(x3)$) {\ldots};
\node at (1.5, 1) {\ldots};

\end{tikzpicture}
\caption{ConvMGU Block}
\label{ConvMGU Block}
\end{figure}
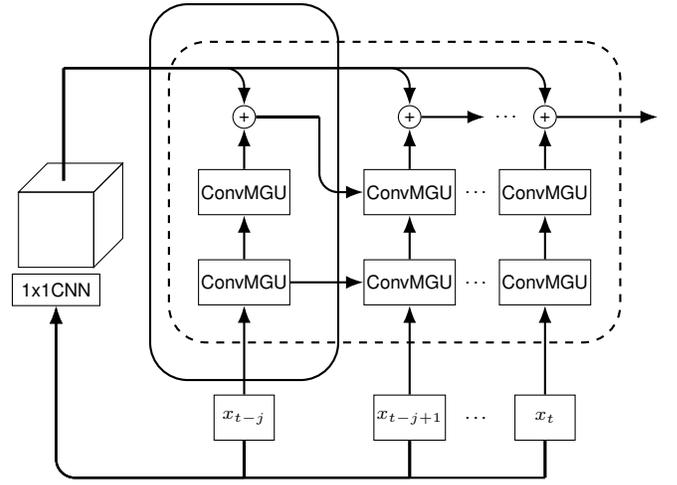

One further modification is shown in Figure \ref{ConvMGU Block}. It stacks two layers of ConvMGU and adds a shortcut later to form a ConvMGU block whose structure is reminiscent of a standard Residual Block\cite{He_2016_CVPR}.

The method proposed by Elsayed \emph{et al}~\cite{DBLP:journals/corr/abs-1810-07251} is used to estimate the number of parameters of the two layer ConvMGU stack, ConvMGU Block and ConvLSTM stack as detailed in Equations 3-5.

\begin{equation}
\centering
\label{key}
	\begin{aligned}
	2\ Layer\ ConvLSTM: 2 \times 4 \times (m^2(\gamma+ \kappa) + 1)\cdot n 
	\end{aligned}
	\end{equation}

\begin{equation}
\centering
\label{key}
	\begin{aligned}
	Our\ Block: 2 \times (m^2(\gamma+ \kappa) + 1)\cdot n + (f^3 \cdot \gamma + 1) \cdot n 
	\end{aligned}
	\end{equation}

\begin{equation}
\centering
\label{key}
	\begin{aligned}
	Our\ 2\ Layer\ ConvMGU: 2 \times (m^2(\gamma+ \kappa) + 1)\cdot n
	\end{aligned}
	\end{equation}

For the purpose of analysis, the assigned values for each hyper parameter is shown in Table \ref{Tab:ConvMGUParam}.

\begin{table}[!h]
	\centering
	\begin{tabular}{|l|l|l|}
		\hline
		& \textbf{Meaning}       & \textbf{Assigned Value} \\ \hline
		\textbf{m}                     & conv kernel size       & 3                       \\ \hline
		$\gamma$ & input channel          & 3                       \\ \hline
		$\kappa$ & number of feature maps & 40                      \\ \hline
		\textbf{n}                     & output channel         & 40                      \\ \hline
		\textbf{f}                     & 3d conv kernel size    & 3                       \\ \hline
	\end{tabular}
	\caption{Unit Setting}
	\label{Tab:ConvMGUParam}
\end{table}

The results demonstrate that the number of parameters of the proposed ConvMGU block is 73\% smaller in comparison with the original two layer ConvLSTM unit.

\subsection{D-Flow Network}

The D-Flow Network is composed of two flows of ConvMGU units that encode spatial temporal features derived from input vision frames. The following convolution layer acts as a simple decoder to perform a one-step prediction of the target area segmentation.

As mentioned previously, handling dynamic lighting conditions is a major problems for mobile robots. The appearance of a colour can vary dramatically when the robot operates in natural lighting conditions. This significantly affects target area segmentation, where the target is defined as a group of pixels that have similar colours. To overcome this problem efficiently, the proposed two-flow network takes both YUV and RGB images as input to efficiently search for the best weights. Figure \ref{fig:1} illustrates the proposed network. 

The whole network performs a single-step prediction that takes a sequence of input frames and attempts to predict the segmentation of the current frame. 
The YUV and RGB images are processed separately by stacks of ConvMGU to extract various spatial temporal features. These features are then added together to produce the final feature maps that the subsequent decoder can use for prediction.

\section{Experiment}

In this section we describe a set experiments to analyse the performance and generalisation ability of our model.

\subsection{SPL Vision Dataset}

For the first experiment, we used a dataset collected from a Nao V5/V6 humanoid robot from Softbank Robotics\cite{lu2021runswift}. The dataset contains 20 video captures where each video contains varying numbers of image sequences with corresponding weak labels. The details of the labelled data are shown in the Table \ref{Dataset Information}. The raw resolution is 1280 by 960. The data were collected from various locations and with different lighting conditions.

\begin{table}[h]
\centering
\resizebox{\columnwidth}{!}{
\begin{tabular}{|l|l|l|l|l|}
\hline
\textbf{\#frame} & \textbf{location} & \textbf{lighting condition} & \textbf{body motion} & \textbf{recording type} \\ \hline
19               & Indoor(Germany)   & Constant(low)            & Static               & Multi-testing           \\ \hline
19               & Indoor(Germany)   & Constant(low)            & Static               & Multi-testing           \\ \hline
19               & Indoor(Germany)   & Constant(low)            & Static               & Multi-testing           \\ \hline
5                & Indoor(Germany)   & Constant(high)           & Static               & Single-testing          \\ \hline
6                & Indoor(Germany)   & Constant(high)           & Static               & Single-testing          \\ \hline
19               & Indoor(Germany)   & Constant(low)            & Dynamic              & Single-testing          \\ \hline
7                & Indoor(Germany)   & Constant(low)            & Dynamic              & Game                    \\ \hline
259              & Indoor(UNSW)      & Constant(medium)         & Dynamic              & Single-testing          \\ \hline
199              & Outdoor(Sydney)   & Night                    & Static               & Single-testing          \\ \hline
121              & Indoor(UNSW)      & Uneven(medium)           & Dynamic              & Single-testing          \\ \hline
271              & Indoor(UNSW)      & Uneven(medium)           & Dynamic              & Multi-testing           \\ \hline
113              & Indoor(UNSW)      & Uneven(medium)           & Dynamic              & Multi-testing           \\ \hline
68               & Outdoor(Sydney)   & Night                    & Static               & Single-testing          \\ \hline
119              & Outdoor(Sydney)   & Night                    & Static               & Single-testing          \\ \hline
298              & Indoor(Sydney)    & Constant(low)            & Dynamic              & Multi-testing           \\ \hline
251              & Outdoor(Sydney)   & Morning(uneven)          & Static               & Single-testing          \\ \hline
210              & Indoor(UNSW)      & Extremely uneven         & Dynamic              & Single-testing          \\ \hline
181              & Indoor(UNSW)      & Extremely uneven         & Dynamic              & Single-testing          \\ \hline
213              & Indoor(UNSW)      & Extremely uneven         & Dynamic              & Multi-testing           \\ \hline
188              & Indoor(UNSW)      & Extremely uneven         & Dynamic              & Multi-testing           \\ \hline
\end{tabular}
}
\caption{Dataset Information}
\label{Dataset Information}
\end{table}

\begin{figure}[!h]
	\centering
	\begin{subfigure}[b]{0.23\textwidth}
		\includegraphics[width=\textwidth]{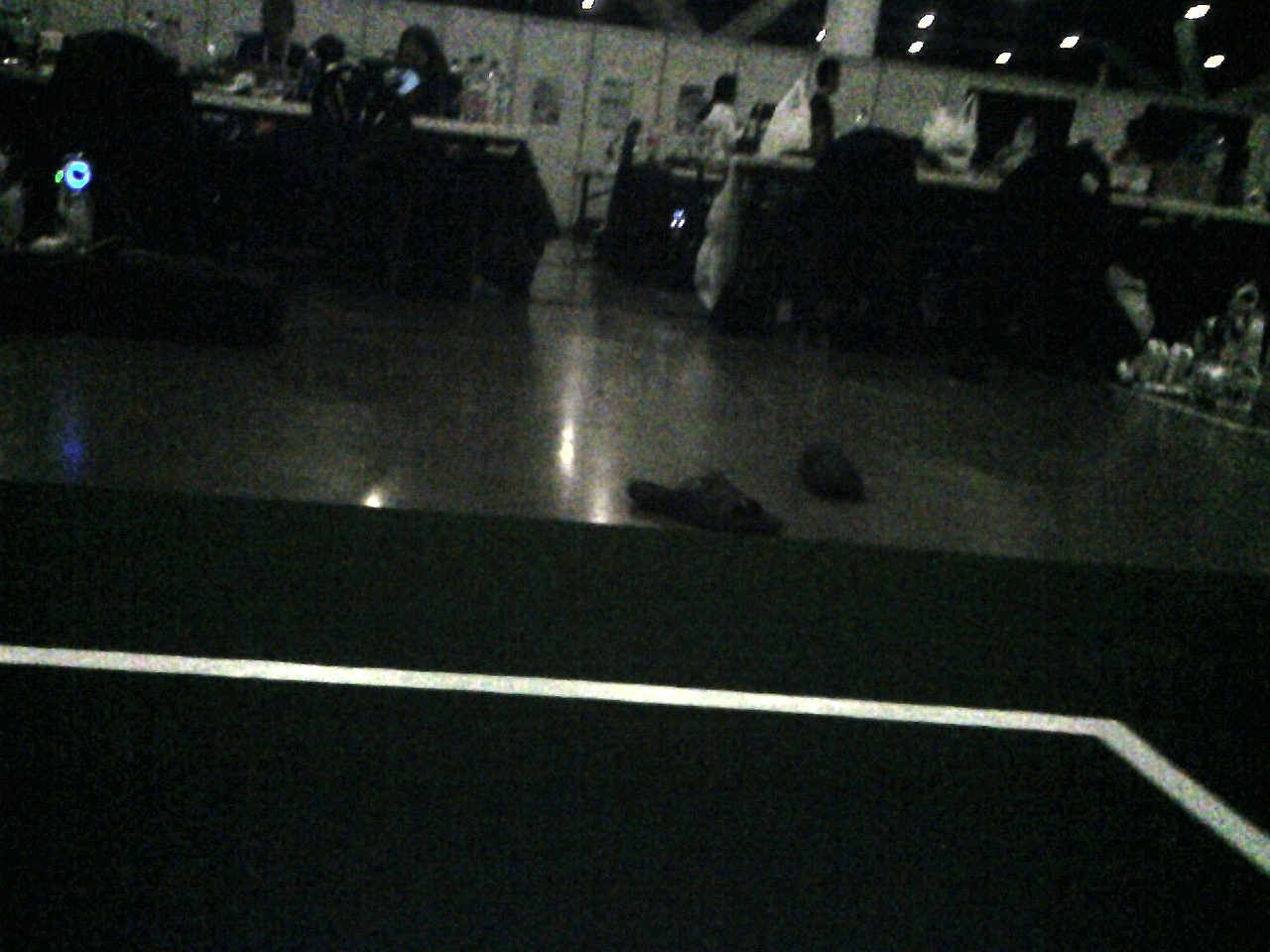}
		\caption{Raw Image}
		\label{fig:splDatasetA}
	\end{subfigure}
	\begin{subfigure}[b]{0.23\textwidth}
		\includegraphics[width=\textwidth]{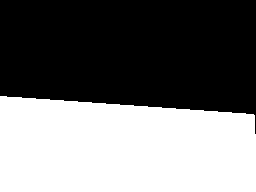}
		\caption{Label}
		\label{fig:splDatasetB}
	\end{subfigure}
	\caption{A sample pair of SPL dataset}
	\label{fig:splDataset}
\end{figure}

Figure \ref{fig:splDataset} presents a pair of a raw images and a label that is used to identify noisy target areas using white shapes. A labelling tool is used to manually mark and generate the white shape that is considered to be the labelled target area. The goal for this dataset is to train a model that can identify these target areas with minimal false positives under dynamic lighting conditions. We hypothesise that such a model may even be able to overcome issues introduced by the weak labelling. Since it is a weak labelling system, the lines and all other objects present within the target area are all given the same label.

\subsection{Experiment Results}

In this experiment we performed target area segmentation with Fast-SCNN, a Fully Convolution Network and our model. Input images were down-sampled four times to 320 by 240 for efficiency. All training was performed on a single NVIDIA RTX 1060 and the inference time analyses were done on the same GPU. The experiment was implemented using the Pytorch 1.7.1~\cite{paszke2017automatic}.

\begin{figure*}[t]
	\centering
	
	\begin{subfigure}[b]{0.15\textwidth}
		\includegraphics[width=\textwidth]{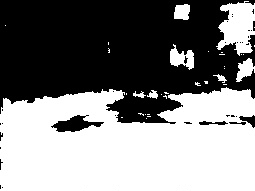}
		\caption{D-Flow\_Small}
		\label{fig:splExperiment1a}
	\end{subfigure}
	\begin{subfigure}[b]{0.15\textwidth}
		\includegraphics[width=\textwidth]{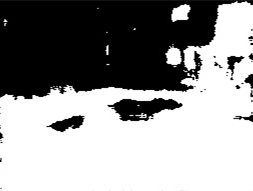}
		\caption{D-Flow}
		\label{fig:splExperiment1b}
	\end{subfigure}
	\begin{subfigure}[b]{0.15\textwidth}
		\includegraphics[width=\textwidth]{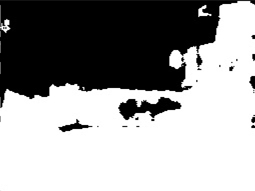}
		\caption{D-Flow\_Focal}
		\label{fig:splExperiment1c}
	\end{subfigure}
	\begin{subfigure}[b]{0.15\textwidth}
		\includegraphics[width=\textwidth]{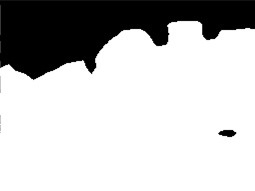}
		\caption{Fast\_SCNN}
		\label{fig:splExperiment1d}
	\end{subfigure}
	\begin{subfigure}[b]{0.15\textwidth}
		\includegraphics[width=\textwidth]{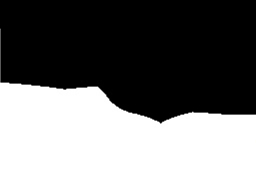}
		\caption{FCN(VGG)}
		\label{fig:splExperiment1e}
	\end{subfigure}
	\begin{subfigure}[b]{0.15\textwidth}
		\includegraphics[width=\textwidth]{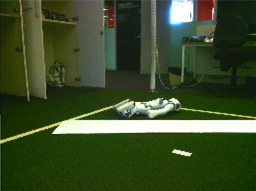}
		\caption{Raw}
		\label{fig:splExperiment1f}
	\end{subfigure}
	\caption{Comparison Experiment Result 1}
	\label{fig:splExperiment1}
\end{figure*}

\begin{figure*}[h]
	\centering
	
	\begin{subfigure}[b]{0.15\textwidth}
		\includegraphics[width=\textwidth]{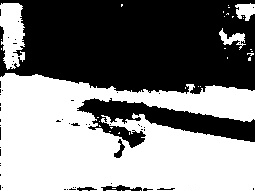}
		\caption{D-Flow\_Small}
		\label{fig:splExperiment2a}
	\end{subfigure}
	\begin{subfigure}[b]{0.15\textwidth}
		\includegraphics[width=\textwidth]{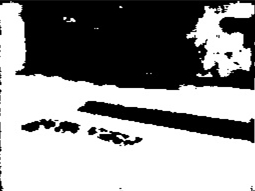}
		\caption{D-Flow}
		\label{fig:splExperiment2b}
	\end{subfigure}
	\begin{subfigure}[b]{0.15\textwidth}
		\includegraphics[width=\textwidth]{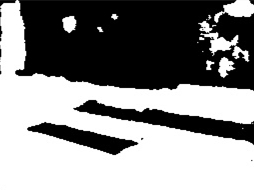}
		\caption{D-Flow\_Focal}
		\label{fig:splExperiment2c}
	\end{subfigure}
	\begin{subfigure}[b]{0.15\textwidth}
		\includegraphics[width=\textwidth]{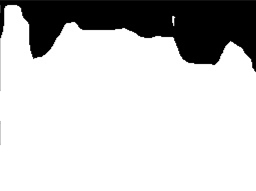}
		\caption{Fast\_SCNN}
		\label{splExperiment2d}
	\end{subfigure}
	\begin{subfigure}[b]{0.15\textwidth}
		\includegraphics[width=\textwidth]{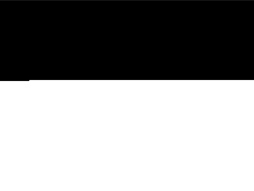}
		\caption{FCN(VGG)}
		\label{fig:splExperiment2e}
	\end{subfigure}
	\begin{subfigure}[b]{0.15\textwidth}
		\includegraphics[width=\textwidth]{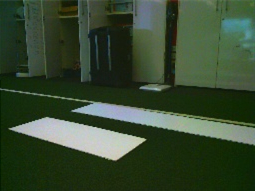}
		\caption{Raw}
		\label{fig:splExperiment2f}
	\end{subfigure}
	\caption{Comparison Experiment Result 2}
	\label{fig:splExperiment2}
\end{figure*}

Figures \ref{fig:splExperiment1} and \ref{fig:splExperiment2} present the qualitative results for two test sequences generated by the candidate methods with different loss functions. The third set of results was generated by our model with Focal Loss~\cite{lin2017focal} while the others were trained with Binary Cross Entropy Loss.

As the images show, the FCN only learned positional patterns and failed to generalise properly in different scenarios. Fast-SCNN learnt some out-of-domain patterns, as it showed the ability to generalise through its identification of the green area from the bottom part of the images. Unfortunately, it tended to overfit based on the colour pattern since it labelled all the green parts as the target area even though some of them were simple light reflections. The left-most three images demonstrate the ability of our model to outperform the other models in this task. All of them label most parts of the target area with minimal outliers that can be easily removed through post-processing. However, only our models were able to crop out random objects in the target area. This  demonstrates that our model can work effectively with a weak labelling system which has noisy labels.

For a quantitative analysis, we used two different metrics to measure segmentation performance. The Dice Coefficient~\cite{doi:10.2307/1932409} is a statistic that measures the similarity between two samples that we use in this experiment to measure the model's ability to learn label patterns. This score only measures how well the model fits the labels, where the labels may be noisy. Therefore, it does not measure the true ability for target area segmentation but acts as a base check for learning outcomes. The silhouette score~\cite{ROUSSEEUW198753} is another metric that is a model-free measurement of the consistency in segmentation. The inference time analysis shown was performed on a single NVIDIA RTX 1060 GPU. The results are shown in Table \ref{Quantitative Results}.

\begin{table}[h]
\centering
\resizebox{\columnwidth}{!}{
	\begin{tabular}{|l|l|l|l|l|}
		\hline
		\textbf{}                              & \textbf{Dice Score} & \textbf{Silhouette score} & \textbf{Inference Time} & \textbf{\#Params} \\ \hline
		\textbf{Ours\_small} & 0.82                      & 26\%                    & 19 ms                   & \textbf{45K}      \\ \hline
		\textbf{Ours}        & 0.83                      & \textbf{27\%}           & 41 ms                   & 0.178M            \\ \hline
		\textbf{FCN(VGG)}                      & 0.85                      & 24\%                    & 31 ms                   & 32M               \\ \hline
		\textbf{Fast-SCNN}                     & \textbf{0.91}             & 26\%                    & \textbf{12ms}           & 1.1M             \\ \hline
	\end{tabular}
	}
	\caption{Quantitative Results}
	\label{Quantitative Results}
\end{table}

%todo add raw data graph, analyze the results

\begin{figure*}[!]
	\centering
	
	\begin{subfigure}[b]{0.15\textwidth}
		\includegraphics[width=\textwidth]{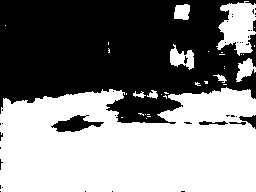}
		\caption{D-Flow\_BCE-Loss}
		\label{fig:handExperiment1a}
	\end{subfigure}
	\begin{subfigure}[b]{0.15\textwidth}
		\includegraphics[width=\textwidth]{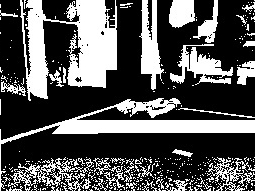}
		\caption{Mean Adaptive-threshold}
		\label{fig:handExperiment1b}
	\end{subfigure}
	\begin{subfigure}[b]{0.15\textwidth}
		\includegraphics[width=\textwidth]{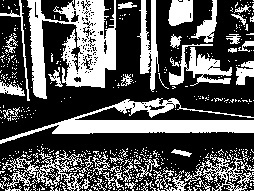}
		\caption{Gaussian Adaptive-threshold}
		\label{fig:handExperiment1c}
	\end{subfigure}
	\begin{subfigure}[b]{0.15\textwidth}
		\includegraphics[width=\textwidth]{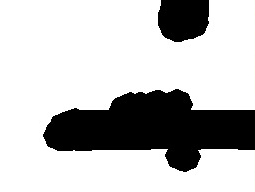}
		\caption{Distance-transform threshold}
		\label{fig:handExperiment1d}
	\end{subfigure}
	\begin{subfigure}[b]{0.15\textwidth}
		\includegraphics[width=\textwidth]{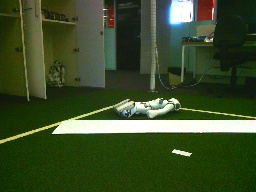}
		\caption{Raw Robot Camera Image}
		\label{fig:handExperiment1e}
	\end{subfigure}
	\caption{Experiment comparing against Handcrafted Methods Result 1}
	\label{fig:handExperiment1}
\end{figure*}

\begin{figure*}[!]
	\centering
	
	\begin{subfigure}[b]{0.15\textwidth}
		\includegraphics[width=\textwidth]{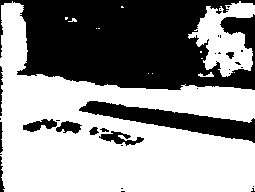}
		\caption{D-Flow\_BCE-Loss}
		\label{fig:handExperiment2a}
	\end{subfigure}
	\begin{subfigure}[b]{0.15\textwidth}
		\includegraphics[width=\textwidth]{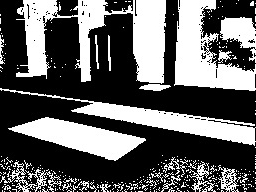}
		\caption{Mean Adaptive-threshold}
		\label{fig:handExperiment2b}
	\end{subfigure}
	\begin{subfigure}[b]{0.15\textwidth}
		\includegraphics[width=\textwidth]{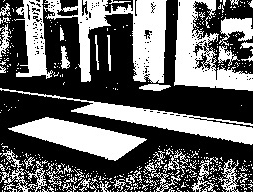}
		\caption{Gaussian Adaptive-threshold}
		\label{fig:handExperiment2c}
	\end{subfigure}
	\begin{subfigure}[b]{0.15\textwidth}
		\includegraphics[width=\textwidth]{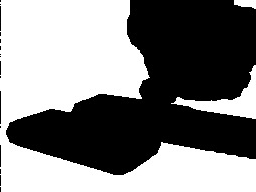}
		\caption{Distance-transform threshold}
		\label{fig:handExperiment2d}
	\end{subfigure}
	\begin{subfigure}[b]{0.15\textwidth}
		\includegraphics[width=\textwidth]{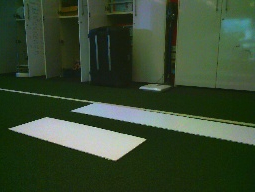}
		\caption{Raw Robot Camera Image}
		\label{fig:handExperiment2e}
	\end{subfigure}
	\caption{Experiment comparing against Handcrafted Methods Result 2}
	\label{fig:handExperiment2}
\end{figure*}
\subsection{Comparison with Handcrafted Methods}

In this section, we compare the proposed model with two handcrafted methods used by the RoboCup team, rUNSWift, where both of them perform field recognition~\cite{brameldrobocup}.

Adaptive-thresholding is an algorithm that groups pixels into binary classes in a sliding window. As shown in Figures \ref{fig:handExperiment1} and \ref{fig:handExperiment2}, two variants are actually doing brightness-based clustering which is suitable for finding certain objects. Unfortunately, these methods are not able to tell the difference between the main part of the soccer field and the image background, where in most cases, they are both shown as a darker colour.

A distance-transform is one step in the Watershed background extraction\cite{roerdink2000watershed}. Since its purpose is to distinguish between the foreground and background, it is not suitable for use in target area segmentation. As shown, it marks white clusters inside the soccer field as well as random bright pixels as foreground. Therefore, its output is a mixture of a part of the target area and a part of the darker background.

\subsection{Experiments on Maritime Dataset}

We tested our work on a maritime dataset to see how well the proposed method performs in different domains. The Singapore Maritime Dataset(SMD)~\cite{7812788} was used in this experiment. It contains two different subsets. In our experiment, we only used the one where the camera is mounted on a moving vessel, which is similar to a mobile  robot.

\begin{table}[h]
\centering
\resizebox{0.9\columnwidth}{!}{
\begin{tabular}{|l|l|l|l|l|}
\hline
\textbf{Dataset} & \textbf{Sub-set} & \textbf{\#Video} & \textbf{\#Frames/Video} & \textbf{\#Total} \\ \hline
SMD              & On-Board         & 11               & 300                     & 3300        \\ \hline   
\end{tabular}
}
\caption{SMD dataset}
\label{Tab:SMD Dataset}
\end{table}

\begin{figure}[h]
	\centering
	
	\begin{subfigure}[b]{0.23\textwidth}
		\includegraphics[width=\textwidth]{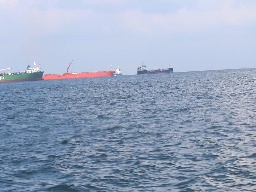}
		\caption{Raw Image}
		\label{fig:maritimeDataset1}
	\end{subfigure}
	\begin{subfigure}[b]{0.23\textwidth}
		\includegraphics[width=\textwidth]{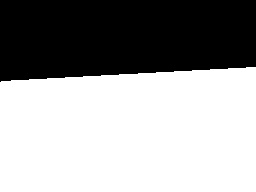}
		\caption{Label}
		\label{fig:maritimeDataset2}
	\end{subfigure}
	
	\caption{a sample pair of SMD dataset}
	\label{fig:maritimeDataset}
\end{figure}

Table \ref{Tab:SMD Dataset} gives the details of the subset of the data we used in this test. We used 9 of the 11 videos as training data and the remaining 2 as test data. In this task, we defined the target area as the ocean below horizon. Figure \ref{fig:maritimeDataset} shows a pair of sample raw frames and the corresponding label. Again, we used a weak labelling system as we only put a rectangular mask as the target.

\begin{figure}[h]
	\centering
	
	\begin{subfigure}[b]{0.23\textwidth}
		\includegraphics[width=\textwidth]{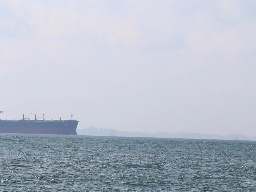}
		\caption{Raw Image}
		\label{fig:maritimeExperimenta}
	\end{subfigure}
	\begin{subfigure}[b]{0.23\textwidth}
		\includegraphics[width=\textwidth]{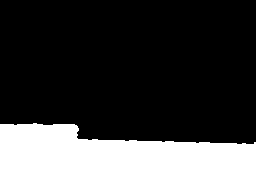}
		\caption{Prediction}
		\label{fig:maritimeExperimentb}
	\end{subfigure}
	\begin{subfigure}[b]{0.23\textwidth}
		\includegraphics[width=\textwidth]{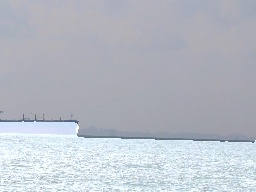}
		\caption{Annotated Image}
		\label{fig:maritimeExperimentc}
	\end{subfigure}
	
	\caption{Sample output for SMD dataset}
	\label{fig:maritimeExperiment}
\end{figure}

Figure~\ref{fig:maritimeExperiment} shows a sample output for the test data. We can see that despite the big difference in the ocean colour between the test data and the sample training data, our model can still find the target region of the ocean below the horizon with not many false positives.

\subsection{Experiment with Different Colour Spaces}

Our model is composed of two sub-nets that process RGB and YUV formatted data separately until the processed feature maps are combined for the final prediction stage.

We conducted a series of experiments with the SMD dataset to understand the effects of using different colour spaces and the network architecture. Table \ref{Tab:Colour Space Experiment details} in Appendix shows the setting for each experiment.

Figure \ref{fig:learningCurve} in the Appendix shows the learning curves for each experiment. We see the RGB-YUV combination with the Double-Flow architecture achieves the most stable performance and lowest validation loss at the 100th epoch. The YUV-only model suffers from underfitting while HSV-only and the HSV-YUV combination suffer from overfitting. The RGB-YUV(Y-Only) experiment is intended to determine if the gray scale image has the same effect as a full YUV image. The RGB-YUV(Y-Only) can achieve low validation loss at some epochs, but the learning curve is quite unstable and tends to fluctuate dramatically while achieving low validation loss. In conclusion, among all these different colour space settings and network architectures, the RGB-YUV combination with Double-Flow net is the best.

\section{Conclusion}

We have proposed a new approach to real time robot vision target area segmentation along with a new parameter reduced convolutional recurrent unit. The proposed D-Flow Network demonstrates the ability to overcome problems including dynamic lighting conditions. We have also created and published a robot vision dataset that was obtained using  a Nao V5/V6 robot. This dataset can be used for a variety of robot vision problems.

\begin{appendices}

\section{Learning Curves for Different Colour Space}
\label{appendix:graph}

\begin{table}[!h]
\centering
\begin{tabular}{|l|l|l|}
\hline
\textbf{Models} & \textbf{Colour for 1st flow} & \textbf{Colour for 2nd flow} \\ \hline
Single-Flow                   & RGB                                & N/A                                 \\ \hline
Single-Flow                   & HSV                                & N/A                                 \\ \hline
Single-Flow                   & YUV                                & N/A                                 \\ \hline
Double-Flow                   & RGB                                & YUV                                 \\ \hline
Double-Flow                   & RGB                                & HSV                                 \\ \hline
Double-Flow                   & HSV                                & YUV                                 \\ \hline
Double-Flow                   & RGB                                & YUV(Y-Only)                         \\ \hline
\end{tabular}
\caption{Experiment Settings}
\label{Tab:Colour Space Experiment details}
\end{table}
\begin{figure}[H]
	\centering
	
	\begin{subfigure}[b]{0.45\textwidth}
		\includegraphics[width=\textwidth]{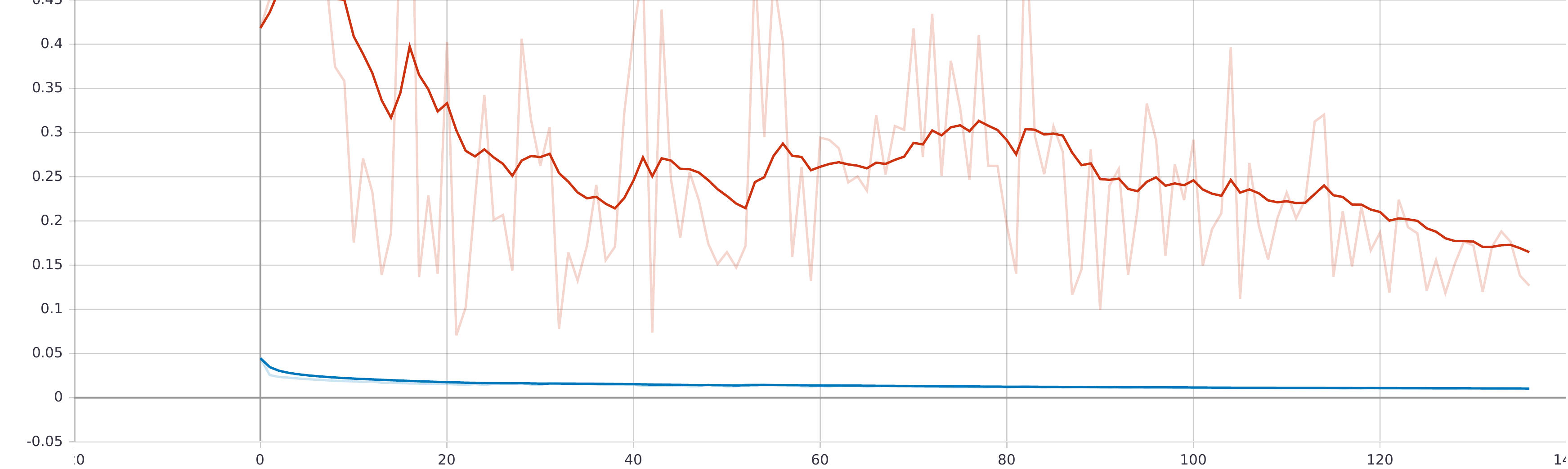}
		\caption{HSV+RGB}
		\label{fig:learningCurvea}
	\end{subfigure}
	\begin{subfigure}[b]{0.45\textwidth}
		\includegraphics[width=\textwidth]{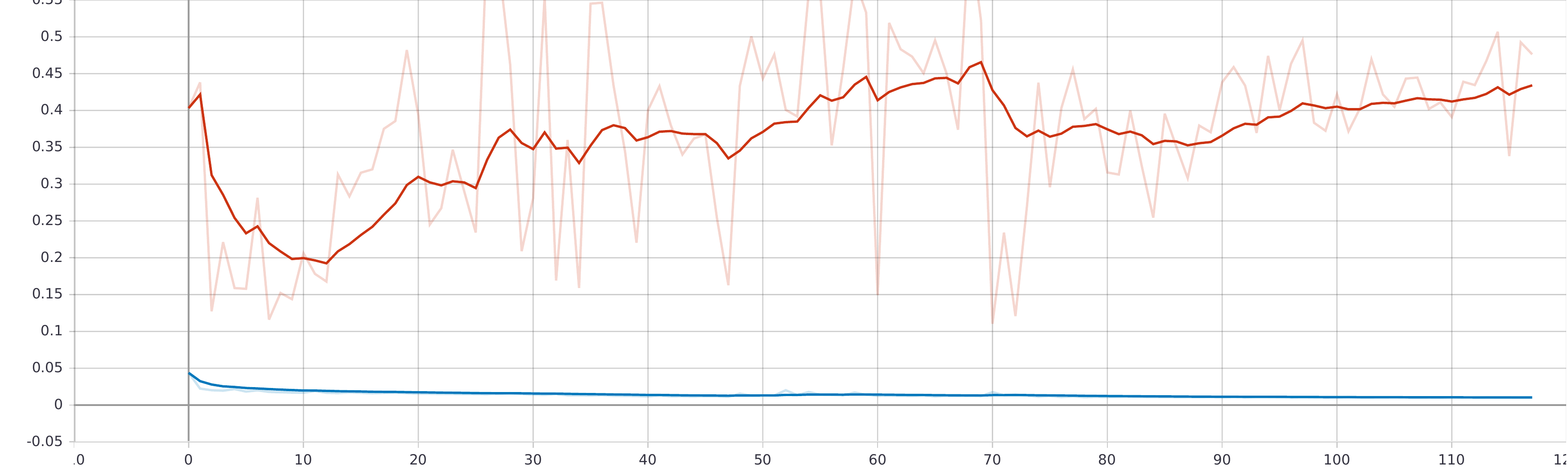}
		\caption{HSV+YUV}
		\label{fig:learningCurveb}
	\end{subfigure}
		\begin{subfigure}[b]{0.45\textwidth}
		\includegraphics[width=\textwidth]{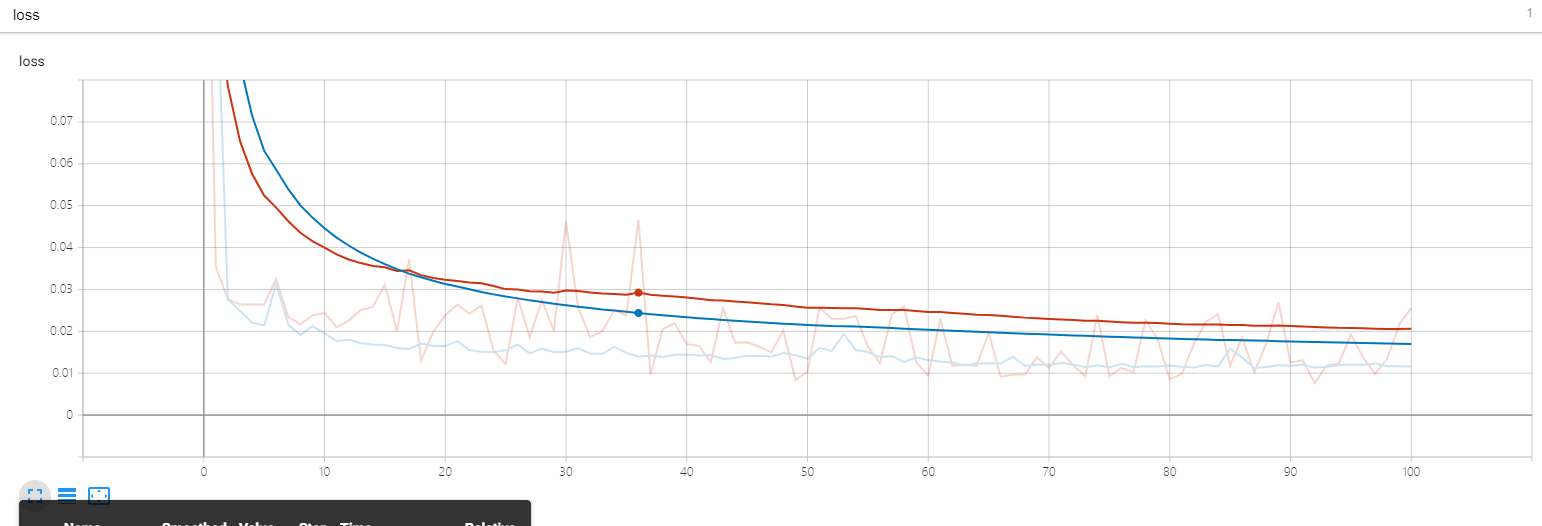}
		\caption{RGB+YUV}
		\label{fig:learningCurvec}
	\end{subfigure}
		\begin{subfigure}[b]{0.45\textwidth}
		\includegraphics[width=\textwidth]{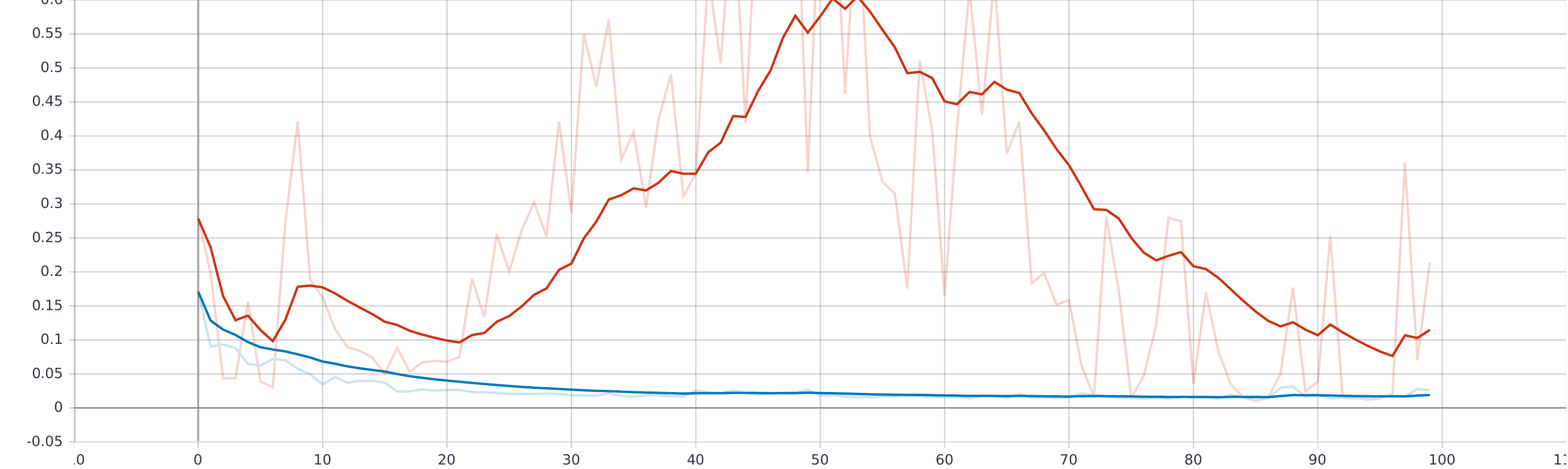}
		\caption{RGB+YUV(Y-Only)}
		\label{fig:learningCurved}
	\end{subfigure}
%\end{figure}
%\begin{figure}[H]
	\begin{subfigure}[b]{0.45\textwidth}
		\includegraphics[width=\textwidth]{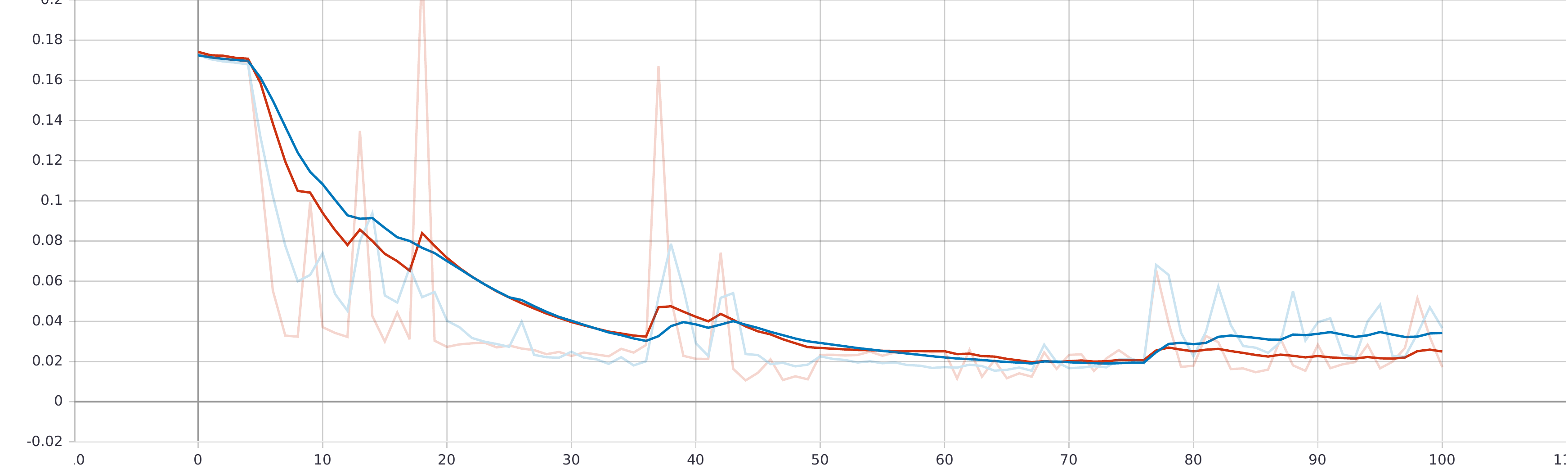}
		\caption{RGB}
		\label{fig:learningCurvee}
	\end{subfigure}
	\begin{subfigure}[b]{0.45\textwidth}
		\includegraphics[width=\textwidth]{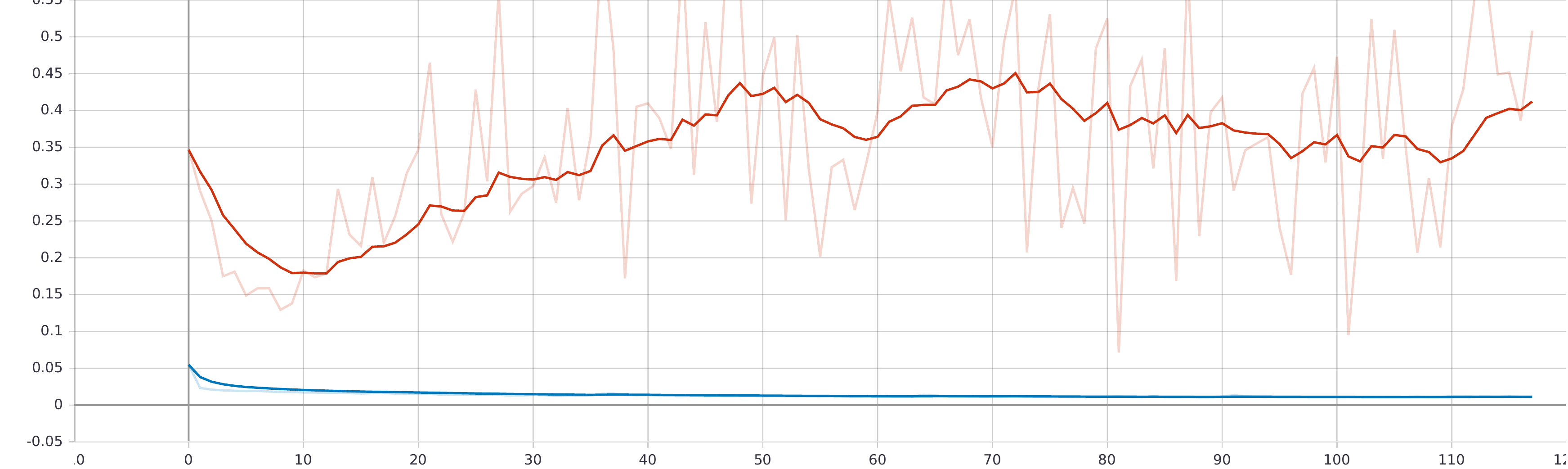}
		\caption{HSV}
		\label{fig:learningCurvef}
	\end{subfigure}
	\begin{subfigure}[b]{0.45\textwidth}
		\includegraphics[width=\textwidth]{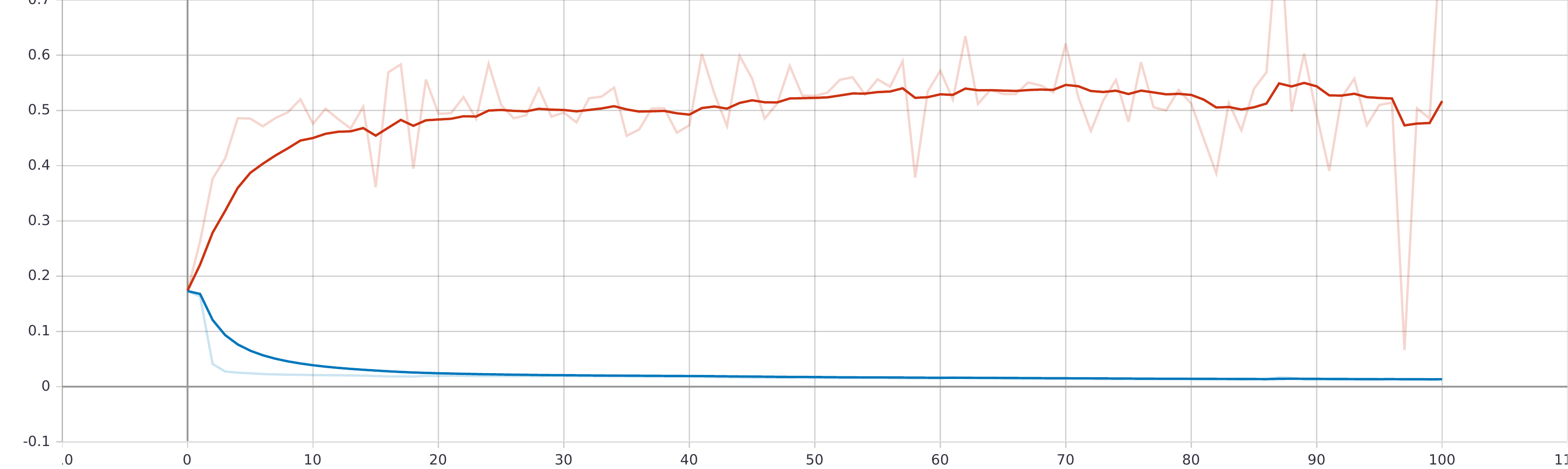}
		\caption{YUV}
		\label{fig:learningCurveg}
	\end{subfigure}
	
	\caption{Learning Curves}
	\label{fig:learningCurve}
\end{figure}

\end{appendices}

\bibliography{aaai22.bib}
\end{document}